\documentclass[journal]{IEEEtran}

\usepackage{amsmath,amssymb,amsfonts,bm}
\usepackage{booktabs}
\usepackage{threeparttable}
\usepackage{makecell}
\usepackage{multirow}
\usepackage{graphicx}
\usepackage{url}
\usepackage[hidelinks]{hyperref}
\usepackage{hyperref}
\usepackage{xcolor}
\newcommand{\mGD}{{\bf GD}}
\newcommand{\mLN}{{\bf LN}}
\newcommand{\mSD}{{\bf SD}}
\newcommand{\mSX}{{\bf SX}}
\graphicspath{{figures/}{figures/chronos2/}{./}}
\allowdisplaybreaks[3]

\title{Market-Information-Aware Gated-LoRA of Foundation Models for Transferable Day-Ahead Electricity Price Forecasting}

\author{Hang Fan,~\IEEEmembership{~Member,~IEEE},
	Wei Wei,~\IEEEmembership{~Senior Member,~IEEE},
    Shengwei Mei,~\IEEEmembership{~Fellow,~IEEE}
	} 

\begin{document}

\maketitle

\begin{abstract}
Electricity price forecasting is crucial for market participants but remains difficult because prices are volatile, market-specific, and closely tied to anticipated system conditions. Existing supervised methods depend largely on market-specific historical data, limiting their use in newly established or data-scarce markets. This paper proposes a market-information-aware adaptation framework that transfers the Chronos-2 time-series foundation model to day-ahead electricity price forecasting. It first constructs a multi-source market information (MSMI) interface aligning 7-day price context with pre-clearing supply--demand, reserve, maintenance, generator-capacity, and intertie variables, and then trains a source-domain gated low-rank adapter (LoRA), updating about $1\%$ of model parameters without target-market labels. The gate scales the frozen source adapter according to reserve-tightness and operating-state signals. A leave-one-market-out protocol is adopted for evaluating cross-market transferability. Experiments on four Chinese provincial day-ahead spot markets show that the proposed framework reduces the average MAE/RMSE by $6.24\%/7.99\%$ relative to market-information-aware zero-shot Chronos-2 and by $3.05\%/3.52\%$ relative to vanilla Source-LoRA. Experiments show that the gain is not reproduced by a learned global scalar or by random gate initialization, while the additional improvement over Source-LoRA is limited. These results suggest that market-structured inputs and state-dependent gated LoRA can provide a practical transfer path for data-scarce electricity markets.

\end{abstract}


\section{Introduction}
\label{sec:introduction}

\IEEEPARstart{W}{ith} the proliferation of electricity markets, accurate price forecasting has become essential for market participants. Unlike purely physical variables, market-clearing prices are jointly shaped by renewable fluctuations, network constraints, strategic behavior, and regulatory rules. As a result, price series often combine calendar regularity with abrupt spikes and regime shifts \cite{weron2014electricity}. Probabilistic forecasting is also increasingly important, as participants require not only expected price trajectories but also assessments of downside risk, scarcity events, and interval reliability \cite{nowotarski2018recent}.

This challenge is amplified when spot-market reform coincides with rapid renewable integration. Rising wind and photovoltaic penetration reshapes supply--demand curves, while new or frequently reformed markets often lack the long, stationary records required for supervised learning. The problem is particularly salient in China, where provincial spot markets differ in rules, generation mix, renewable penetration, and operating conditions \cite{guo2020chinamarket}. For such markets, the central question is whether forecasting capability can be transferred to a target market before sufficient local labels become available.

\subsection{Related Work}

Electricity price forecasting has long relied on statistical, machine-learning, and deep-learning models. Reviews show that performance is highly sensitive to market selection, data splitting, and benchmark design \cite{weron2014electricity}, while the open-access benchmark in \cite{lago2021forecasting} highlights the importance of rolling evaluation and strong baselines. High-dimensional autoregressive and regularized models remain competitive when calendar effects and lagged prices are properly encoded \cite{ziel2018day}. Exogenous variables can further improve accuracy, but indiscriminate feature inclusion may weaken generalization \cite{ludwig2015lassoepf}. Given the spikes and heteroscedasticity of spot prices, variance-stabilizing target transformations can also materially affect forecast accuracy \cite{uniejewski2018vst}. Deep-learning approaches include recurrent, attention-based, and hybrid CNN--LSTM models \cite{bottieau2023interpretable, jahangir2020rough, vahedi2026hybrid}, as well as the Temporal Fusion Transformer \cite{lim2021temporal}, Autoformer \cite{wu2021autoformer}, and FEDformer \cite{zhou2022fedformer}. Probabilistic forecasting has likewise advanced through prediction-interval optimization, ensemble density estimation, and generative real-time price models \cite{zhangfu2024probabilistic, chai2019density, zhang2022ganlmp}. Despite improved within-market accuracy, most studies still require substantial target-market data for supervised training and offer limited transferability.

Transfer learning has been explored when target-domain labels are limited. For electricity price forecasting, domain mismatch remains a major obstacle because price dynamics are market-specific \cite{gunduz2023transfer}. In renewable forecasting, multi-domain adaptation has been used to improve probabilistic wind-power prediction under limited target data \cite{dong2023transfer}. These studies motivate the leave-one-market-out design adopted in this paper. A more general transfer mechanism is offered by time-series foundation models, which learn generic forecasting representations from large heterogeneous corpora. Chronos-2 tokenizes numerical sequences for pretrained forecasting \cite{ansari2025chronos2}; Lag-Llama studies probabilistic autoregressive foundation modeling \cite{rasul2023lagllama}; Moirai uses a unified architecture for universal forecasting across different variate structures \cite{woo2024moirai}; and TimesFM studies decoder-only zero-shot forecasting at scale \cite{das2024timesfm}. Recent exogenous-variable Transformers such as TimeXer also suggest that known covariates should be treated as part of the forecasting interface rather than as an afterthought \cite{wang2024timexer}. 

Most recently, PriceFM pretrains a probabilistic electricity-price foundation model across European markets. It focuses on large-scale cross-market pretraining and leaves the adaptation of general-purpose backbones and target-market-free market-information interfaces less explored \cite{yu2025pricefm}. Energy-domain study adapts time-series foundation models to wind-power forecasting with limited target data, addressing heterogeneous source domains and cross-scenario generalization \cite{yan2025windltsm}. This work highlights window construction and mixed cross-scenario training as key to adapting generic time-series models to energy applications.

\begin{table*}[!t]
\centering
\caption{Comparison of representative methods relevant to transferable day-ahead electricity price forecasting.}
\label{tab:related_work}
\begin{threeparttable}
\begin{tabular}{lllccc}
\toprule
Reference & Target & Model family & Future info & Cross-market & Probabilistic \\
\midrule
Ziel and Weron \cite{ziel2018day} & Electricity price & High-dimensional ARX & Yes & No & No \\
Ludwig \emph{et al.} \cite{ludwig2015lassoepf} & Electricity price & LASSO / random forest & Yes & No & No \\
Bottieau \emph{et al.} \cite{bottieau2023interpretable} & Electricity price & Interpretable Transformer & Limited & No & Yes \\
Lago \emph{et al.} \cite{lago2021forecasting} & Electricity price & DNN benchmark & Yes & No & No \\
Lim \emph{et al.} (TFT) \cite{lim2021temporal} & General time series & Attention with covariates & Yes & No & Yes \\
Autoformer / FEDformer \cite{wu2021autoformer,zhou2022fedformer} & General time series & Decomposition Transformer & No & No & No \\
G\"und\"uz \emph{et al.} \cite{gunduz2023transfer} & Electricity price & Transfer learning & Limited & Within-EU & No \\
Chronos \cite{ansari2024chronos} & General time series & Foundation (tokenized) & No & Zero-shot & Yes \\
Lag-Llama \cite{rasul2023lagllama} & General time series & Foundation (autoregressive) & No & Zero-shot & Yes \\
Moirai \cite{woo2024moirai} & General time series & Foundation (universal) & Limited & Zero-shot & Yes \\
TimesFM \cite{das2024timesfm} & General time series & Foundation (decoder-only) & No & Zero-shot & No \\
TimeXer \cite{wang2024timexer} & General time series & Exogenous Transformer & Yes & No & No \\
Yan \emph{et al.} \cite{yan2025windltsm} & Wind power & Foundation + few-shot / re-pretrain & Limited & Cross-farm & Yes \\
\textbf{This work} & \textbf{Electricity price} & \textbf{Foundation + source-domain LoRA} & \textbf{Yes} & \textbf{LOMO} & \textbf{Yes} \\
\bottomrule
\end{tabular}
\begin{tablenotes}
\footnotesize
\item {\bf Future info}: access to known future information over the forecast horizon; {\bf Cross-market}: deployment on a target market without target-market training labels; {\bf Probabilistic}: native distributional forecasts. {\bf ARX}: AutoRegressive with exogenous input; {\bf LASSO}: least absolute shrinkage and selection operator; {\bf DNN}: deep neural network.
\end{tablenotes}
\end{threeparttable}
\end{table*}

\subsection{Research Gap and Contribution}

Table~\ref{tab:related_work} summarizes representative methods along three axes most relevant to the forecasting problem: whether known future information over the forecast horizon is exposed to the model, whether cross-market transfer is supported without using target-market training labels, and whether the model produces probabilistic forecasts. Two observations stand out. First, price forecasting specific studies and exogenous-variable Transformers do leverage external information, but they assume target-market supervision; conversely, time-series foundation models support zero-shot transfer but typically operate as univariate or weakly exogenous-information forecasters. Second, time-series large model studies in energy have so far concentrated on physical processes such as wind power; day-ahead electricity prices have not yet been examined under a target-market-free transfer protocol with full future market-information exposure.

Day-ahead electricity prices expose a gap between generic time-series foundation models and market-aware forecasting. Day-ahead prices are formed by mapping anticipated supply–demand conditions into clearing prices. Therefore, future load, renewable output, reserve conditions, maintenance capacity, generator availability, and intertie schedules available before clearing are not auxiliary variables in a generic sense; they are part of the economic information set from which prices are formed. A price-only foundation-model infers future prices from price history, even when tomorrow's system conditions are already available. Conversely, a fully supervised target market model can use these variables but loses the ability to represent a newly established or label-scarce target market. The missing piece, visible as the unfilled combination in Table~\ref{tab:related_work}, is a transfer framework that excludes target-market labels from training while enabling the foundation model to exploit the information of day-ahead market clearing.

This paper develops a transferable day-ahead electricity price forecasting framework built on Chronos-2, formulating cross-market forecasting as an interface-and-adaptation problem. The main contributions are as follows.
\begin{itemize}
\item \textbf{Multi-source market-information (MSMI) interface.} The MSMI interface provides the pretrained backbone with a 7-day price context, historical covariates, known future information, and day-ahead market-clearing information over the forecasting horizon. Ablation tests demonstrate that this task interface is the primary source of performance improvement.

\item \textbf{Source-domain and gated LoRA adaptation.} We train a lightweight LoRA on source markets and then introduce a market-state gate that scales the LoRA update according to reserve tightness, net load, renewable share, capacity tightness, and recent volatility. This keeps the pretrained backbone frozen while turning the source adapter into a compact operating-state-aware calibration mechanism.

\item \textbf{Target-market-free transfer evidence.} We reformulate day-ahead electricity price forecasting from independent per-market supervised learning to cross-market transfer and evaluate it using a leave-one-market-out protocol. Experiments show that most transferable gains come from market information and source-domain alignment, and the market-state gate cannot be replaced with a learnable scalar or random gate initialization.

\end{itemize}

The proposed method is detailed in Section~\ref{sec:method}. Section~\ref{sec:experiments} describes the experimental setup, main results and limitations. Section~\ref{sec:conclusion} concludes the paper.

\section{Proposed Method}
\label{sec:method}

The day-ahead electricity price forecasting problem is first formulated, followed by an overview of the entire framework. Details are presented in the subsequent subsections.

\subsection{Problem Formulation and Overview}

Let $\mathcal{M}$ denote a set of provincial spot markets. For each market $m \in \mathcal{M}$, we observe a day-ahead price sequence $y_t^{(m)}$ and market-clearing information $\mathbf{x}_t^{(m)}$. Given a historical context window of length $L$, the goal is to forecast the next $H$ day-ahead prices:
\begin{equation}
\hat{\mathbf{y}}_{t+1:t+H}^{(m)} = f\left(\mathbf{x}_{t+1:t+H}^{(m)}|
    \mathbf{y}_{t-L+1:t}^{(m)}, \mathbf{x}_{t-L+1:t}^{(m)}\right).
\end{equation}
In this paper, $L=672$ and $H=96$, corresponding to a 7-day history and a 1-day horizon at a 15-minute resolution.

Chronos-2 serves as a general pretrained time-series forecaster that takes recent target observations and forecast-horizon covariates as inputs and produces probabilistic forecasts with quantile outputs. Its encoder-only architecture directly accommodates historical prices and known future market information without modifying the backbone. This capability is well suited to day-ahead electricity markets, where load, renewable generation, reserves, maintenance, available capacity, and tie-line flow are known or scheduled before clearing, while price spikes and tail risks remain decision-relevant. The resulting setup isolates the transferable value of the market-information interface and source-domain LoRA adaptation from that of architecture redesign.

\begin{figure}[!t]
\centering
\includegraphics[width=0.48\textwidth]{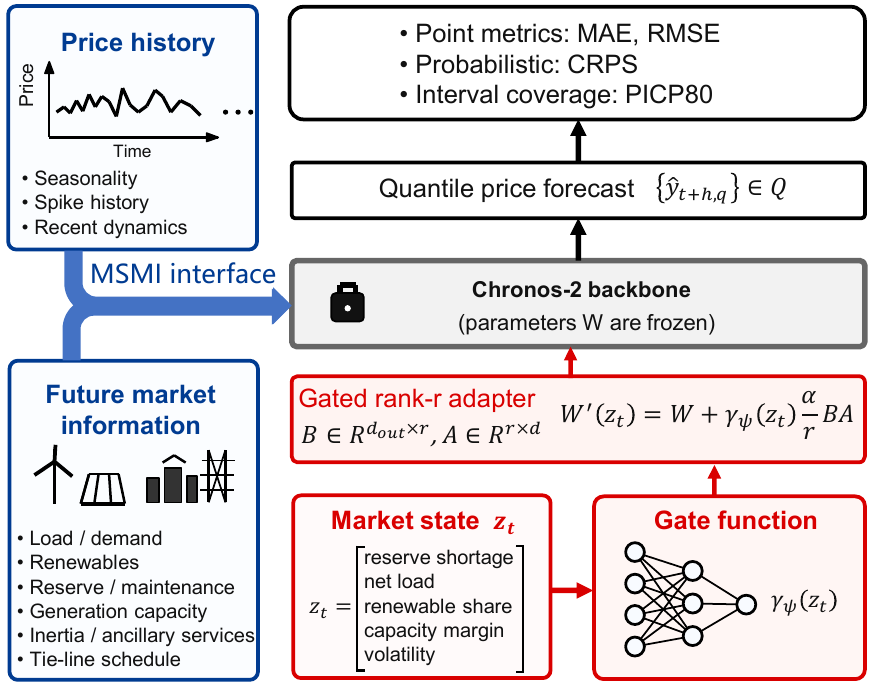}
\caption{An overview of the proposed method}
\label{fig:chronos2_pipeline}
\end{figure}

We evaluate the model under a LOMO protocol. In each fold, one market is held out as the target market. The zero-shot setting evaluates Chronos-2 directly on the held-out market without any training on the dataset. The adaptation setting fine-tunes Chronos-2 on the source markets only and then evaluates it on the held-out market. No target-market training data are used in either zero-shot or source-market adaptation. Fig.~\ref{fig:chronos2_pipeline} summarizes the proposed adaptation framework. The method is deliberately \emph{interface-first}: we do not change the pretrained Chronos-2 backbone. Market knowledge enters through a target-market-free evaluation protocol, a task interface that utilizes day-ahead clearing information, and a low-rank source-domain adapter trained only on non-target markets.

\subsection{Leave-One-Market-Out Task Generation}

Let $m^\star$ denote the held-out target market in one LOMO fold, and $\mathcal{S}(m^\star)=\mathcal{M}\setminus\{m^\star\}$ the source set. Adapter training and testing windows are generated from $\mathcal{S}(m^\star)$ and $m^\star$, respectively, thus target-market labels are not exposed.

For each market $m$ and forecast origin $t$, let $y_t^{(m)}$ be the day-ahead clearing price. In the Core interface, the day-ahead market-information vector
$\mathbf{x}_t^{(m)}\in\mathbb{R}^{d}$ contains load, wind generation, and photovoltaic generation. In the proposed MSMI interface, it is expanded to include additional supply, demand, reserve, maintenance, generator-capacity, and intertie-related fields. A rolling task is defined as
\begin{equation}
    \mathcal{T}_t^{(m)}
    =
    \left(
    \mathbf{y}_{t-L+1:t}^{(m)},
    \mathbf{x}_{t-L+1:t}^{(m)},
    \mathbf{x}_{t+1:t+H}^{(m)};
    \mathbf{y}_{t+1:t+H}^{(m)}
    \right),
\end{equation}
where $L=672$ and $H=96$, corresponding to a seven-day context and a one-day horizon at 15-minute resolution. The source-domain adaptation set in fold $m^\star$ is
\begin{equation}
    \mathcal{D}_{\mathrm{src}}(m^\star)
    =
    \{\mathcal{T}_t^{(m)}:m\in\mathcal{S}(m^\star),\, t\in\Omega_m^{\mathrm{train}}\},
\end{equation}
and the target evaluation set is
\begin{equation}
    \mathcal{D}_{\mathrm{tar}}(m^\star)
    =
    \{\mathcal{T}_t^{(m^\star)}:t\in\Omega_{m^\star}^{\mathrm{test}}\}.
\end{equation}
This construction mimics deployment to a market with no usable local training history. It keeps the rolling-window information set identical for zero-shot, source-LoRA, and later few-shot diagnostics, so performance differences can be attributed to adaptation rather than to different data access.

\subsection{Market-Information-Aware Probabilistic Interface}

The interface passed to Chronos-2 removes the future labels from $\mathcal{T}_t^{(m)}$ and keeps only the information available at the forecast origin:
\begin{equation}
    \mathcal{I}_t^{(m)}
    =
    \left(
    \mathbf{y}_{t-L+1:t}^{(m)},
    \mathbf{x}_{t-L+1:t}^{(m)},
    \mathbf{x}_{t+1:t+H}^{(m)}
    \right).
\end{equation}
The first component provides recent price dynamics, including weekly seasonality and spike history. The second component aligns past prices with contemporaneous system conditions. The third component supplies the day-ahead information vector over the prediction horizon. These quantities are published or scheduled before market clearing. They represent the market participant's information set.

The MSMI interface is designed to be richer than the Core load-wind-PV interface while avoiding hand-crafted transformations as the main source of improvement. It includes direct load, outgoing load, wind power, photovoltaic power, hydro power, nuclear power, upward and downward reserve-related quantities, ancillary-service quantities, maintenance capacity, generation capacity, and tie-line schedule. This design tests whether Chronos-2 benefits from a broader representation of market-clearing conditions before adding model-side complexity. This construction is related to recent exogenous-variable time-series models, but the motivation is different. Architectures such as TimeXer redesign Transformer attention to incorporate external variables during supervised training \cite{wang2024timexer}. Here, the Chronos-2 backbone is kept fixed in the zero-shot setting; we aim to enable a pretrained foundation model to receive the same available information through its task interface while preserving target-market separation.

Chronos-2 returns quantile forecasts instead of a deterministic trajectory. Let $\mathcal{Q}$ denote the evaluated quantile levels and let $\hat{y}_{t+h,q}^{(m)}$ be the predicted $q$-quantile for horizon step $h$. The point forecast is the median quantile $\hat{y}_{t+h,0.5}^{(m)}$. For probabilistic evaluation, we consider the standard pinball loss \cite{nowotarski2018recent}:
\begin{equation}
\rho_q(e_{t+h,q}^{(m)}) =
\max\left(qe_{t+h,q}^{(m)}, (q-1)e_{t+h,q}^{(m)}\right),
\end{equation}
where $e_{t+h,q}^{(m)}=y_{t+h}^{(m)}-\hat{y}_{t+h,q}^{(m)}$ and the continuous ranked probability score (CRPS) \cite{gneiting2007proper}, which can be expressed as twice the integral of the pinball loss over all quantile levels, yielding the following discrete approximation over uniformly spaced quantile grid:
\begin{equation}
\label{eq:Appr-CRPS}
\widehat{\mathrm{CRPS}}_{t+h}^{(m)} = \frac{2}{|\mathcal{Q}|}
\sum_{q\in\mathcal{Q}} \rho_q(e_{t+h,q}^{(m)}).
\end{equation}
As users may need to know whether predicted risk bands contain the realized price with the intended frequency, we also report the prediction interval coverage probability (PICP) for the central $80\%$ interval:
\begin{equation}
\mathrm{PICP}_{80} =
\frac{\sum\limits_{t\in\Omega}\sum\limits_{h=1}^{H}
\mathbb{I}\left(\hat{y}_{t+h,0.1}^{(m)} \le y_{t+h}^{(m)}
\le \hat{y}_{t+h,0.9}^{(m)} \right)}{|\Omega|H}
\end{equation}
PICP values close to 80\% indicate well-calibrated central prediction intervals. 
This probabilistic interface is important for price forecasting because price spikes and tail events are decision-relevant. It allows a single Chronos-2 run to support both median-based point comparisons and distribution-aware evaluation.
For a held-out market $m$, let $\mathrm{PICP}_{80}^{(m)}$ denote
the quantity in the above equation. We report two complementary
cross-market summaries. The pooled coverage,
$\mathrm{PICP}_{80}^{\mathrm{pool}}$, is computed after concatenating
all forecast points from the four held-out markets. To measure
market-wise calibration stability, we additionally report the
unweighted macro-average absolute coverage deviation
\begin{equation}
\label{eq:macro_coverage_deviation}
\mathrm{Macro\text{-}ACD}_{80}
=
\frac{1}{|\mathcal{M}|}
\sum_{m\in\mathcal{M}}
\left|
\mathrm{PICP}_{80}^{(m)} - 0.80
\right|.
\end{equation}
Thus, $\mathrm{PICP}_{80}^{\mathrm{pool}}$ measures aggregate
coverage, whereas $\mathrm{Macro\text{-}ACD}_{80}$ detects
under- or over-coverage that may be masked by aggregation across
markets.

\subsection{Source-Domain Gated LoRA}

The adaptation strategy contains three deployment regimes. In the first regime, zero-shot inference, the pretrained Chronos-2 parameters $\theta_0$ are frozen and each target-market input $\mathcal{I}_t^{(m^\star)}$ is evaluated directly. This is the strictest deployment condition and requires no market-specific training.

In the second regime, source-domain LoRA adaptation, Chronos-2 is adapted using only $\mathcal{D}_{\mathrm{src}}(m^\star)$. For an adapted weight matrix $W$, LoRA represents the update as \cite{hu2022lora}
\begin{equation}
    W' = W + \frac{\alpha}{r}BA,
\end{equation}
where $r$ is the adapter rank, $W \in \mathbb{R}^{d_{\text {out }} \times d_{\text {in }}}$, $A \in \mathbb{R}^{r \times d_{\text {in}}}$, $B \in \mathbb{R}^{d_{\text{out}} \times r}$. Pretrained weights $\theta_0$ remain frozen. The adapter parameters $\phi$ are optimized on source-market rolling tasks:
\begin{equation}
    \begin{split}
    \phi^\star(m^\star)
    =
    \arg\min_{\phi}
    \sum_{\mathcal{T}\in\mathcal{D}_{\mathrm{src}}(m^\star)}
    \mathcal{L}_{\mathrm{C2}}
    \left(
    \mathcal{I}_{\mathcal{T}},
    \mathbf{y}_{\mathcal{T}};
    \theta_0,\phi
    \right),
    \end{split}
\end{equation}
where $\mathcal{L}_{\mathrm{C2}}$ denotes the sequence-forecasting objective used by Chronos-2 for each rolling task. Unless otherwise stated, the MSMI source-adaptation experiments use a controlled LoRA configuration with rank $r=8$, scaling factor $\alpha=16$, 1500 optimization steps, batch size 8, learning rate $10^{-4}$, and bfloat16 arithmetic. The low-rank updates are restricted to the four self-attention projection matrices that form the query, key, value, and output transformations, together with the output projection that maps patch embeddings to forecast tokens. This setting updates approximately 1.21M adapter parameters in each fold, or about 1.0\% of the 120.7M total model parameters, while all pretrained backbone weights remain frozen. After source-domain optimization, the resulting adapter is transferred unchanged to $\mathcal{D}_{\mathrm{tar}}(m^\star)$ for held-out-market evaluation, with no target-market labels used for training or model selection.

The vanilla LoRA applies the same adapter strength to all operating conditions. This may not be appropriate for electricity prices, because the usefulness of an adapter depends on market conditions. To this end, we introduce a market-state-gated LoRA mechanism. For each rolling task, a compact market-state vector is constructed from the known day-ahead market-information trajectory and the recent price context. Let $\ell_{t+h}$, $w_{t+h}$, $p_{t+h}$, $r^{\mathrm{up}}_{t+h}$, $r^{\mathrm{down}}_{t+h}$, $c^{\mathrm{mnt}}_{t+h}$, $g^A_{t+h}$, and $g^B_{t+h}$ denote day-ahead load, wind generation, photovoltaic generation, upward reserve, downward reserve, maintenance capacity, and two available-generation-capacity fields for horizon slot $h$. With $\bar{\ell}_t=H^{-1}\sum_{h=1}^{H}\ell_{t+h}$ and $d_t=\max(|\bar{\ell}_t|,1)$, the raw state scores are
\begin{align}
s_t^{\mathrm{netload}} &= \frac{1}{H}\sum_{h=1}^{H}\frac{\ell_{t+h}-w_{t+h}-p_{t+h}}{d_t}, \nonumber\\
s_t^{\mathrm{renewable}} &= \frac{1}{H}\sum_{h=1}^{H}\frac{w_{t+h}+p_{t+h}}{d_t}, \nonumber\\
s_t^{\mathrm{resAdeq}} &= \frac{1}{H}\sum_{h=1}^{H}\frac{r^{\mathrm{up}}_{t+h}+r^{\mathrm{down}}_{t+h}}{d_t}, \nonumber\\
s_t^{\mathrm{capacity}} &= \frac{1}{H}\sum_{h=1}^{H}
\frac{g^A_{t+h}+g^B_{t+h}-c^{\mathrm{mnt}}_{t+h}-\bar{\ell}_t}{d_t}, \nonumber\\
s_t^{\mathrm{volatility}} &= \frac{\mathrm{std}(y_{t-L+1:t})}{\mathrm{mean}(|y_{t-L+1:t}|)+10^{-6}}  \nonumber
\end{align}
The standardized market-state vector is then
\begin{equation}
    \mathbf{z}_t = [
    \tilde{s}_t^{\mathrm{netload}},
    \tilde{s}_t^{\mathrm{renewable}},
    -\tilde{s}_t^{\mathrm{resAdeq}},
    -\tilde{s}_t^{\mathrm{capacity}},
    \tilde{s}_t^{\mathrm{volatility}}] \notag
\end{equation}
The third component is a reserve-tightness score. The negative signs convert high available reserve and high capacity margin into low tightness scores, so larger positive components indicate tighter operating states. In each LOMO fold, the mean and standard deviation used for standardization are estimated only from source-market training rolling tasks. Held-out target forecast labels and target-test distribution statistics are never used in this scaler. Raw scores are clipped to $[-5,5]$, and if fewer than half of the horizon load entries are valid, $\mathbf{z}_t$ is set to zero. A small gate $g_\psi(\cdot)$ maps $\mathbf{z}_t$ to a task-level LoRA multiplier:
\begin{equation}
\gamma_\psi(\mathbf{z}_t) = \mathrm{clip} \left(   1 + \beta \tanh(g_\psi(\mathbf{z}_t)),  \gamma_{\min},  \gamma_{\max} \right). 
\end{equation}
The adapted layer then becomes
\begin{equation}
\label{eq:dynamic_lora_layer}
    W'(\mathbf{z}_t) = W+\gamma_\psi(\mathbf{z}_t)\frac{\alpha}{r}BA.
\end{equation}
In this gated LoRA mechanism, $\gamma_\psi(\mathbf{z}_t)$ adaptively modulates the adapter strength according to real-time market conditions. It is computed once for each rolling day-ahead task and is applied to the LoRA update during the Chronos-2 forward pass. Thus, the median and quantile forecasts are generated by the gated backbone itself, not by rescaling the final price forecast. In the most stable implementation, the gate is initialized from the source-domain LoRA checkpoint. The LoRA matrices $A$ and $B$ are then frozen, and only the gate parameters $\psi$ are trained on source-market rolling tasks:
\begin{equation}
    \psi^\star(m^\star)  = \arg\min_{\psi}
    \sum_{\mathcal{T}\in\mathcal{D}_{\mathrm{src}}(m^\star)}
    \mathcal{L}_{\mathrm{C2}}\left( \mathcal{I}_{\mathcal{T}},
    \mathbf{y}_{\mathcal{T}};    \theta_0,\phi^\star,\psi\right).  \notag
\end{equation}
This source-initialized gate-only design preserves the transferable adapter learned from related markets and adds only a small market-state calibration layer. It is also deliberately conservative: the gated module cannot memorize the held-out target market, because its trainable gate sees only source-market rolling tasks during adaptation.

Target-market few-shot refinement is used only as a deployment diagnostic. For each held-out market, Chronos-2 is adapted with 5, 10, or 30 days of target-market observations and then evaluated on the same rolling test windows. This tests whether local specialization is useful immediately or requires source-domain alignment first. In practice, the proposed deployment path is: market-information-aware zero-shot inference, source-domain LoRA when related-market data are available, gated market-state calibration of the source adapter, and optional few-shot target refinement after local observations accumulate.

Taken together, the three regimes form a progressive deployment spectrum. A market participant can begin with market-information-aware zero-shot inference, add source-domain LoRA when related-market data are available, and apply target-market few-shot refinement only after adequate local history has accumulated. This staged design is the main methodological distinction from full domain re-pretraining: instead of building a new electricity-price foundation model from a massive cross-market corpus, we adapt a pretrained model through a small source-market adapter while preserving a clean target-market separation.

\section{Experimental Study}
\label{sec:experiments}

This section reports the experimental details and results, and also discusses the limitations of the proposed method.

\subsection{Experimental Setup}

\subsubsection{Dataset, version policy, and protocol}

The raw data consist of two tables: a price table containing day-ahead and real-time clearing prices, and a supply-demand table containing load, renewable generation, reserve, maintenance, generation capacity, and intertie variables. We retain four comparable provincial day-ahead spot markets and exclude non-comparable records from the main LOMO experiment. The data are sampled at 15-minute resolution. For every market, the last 90 days are used for rolling evaluation, and the preceding period, capped at the most recent 900 days per market, is used as the source-domain training pool when that market is not held out. Two markets provide longer histories than this cap; Section~\ref{sec:results} verifies that extending the training pool to the full available history does not change the conclusions.

A key pre-processing issue is that the supply-demand table can contain multiple versions for the same market, date, and 15-minute slot. To avoid hindsight leakage, we select one row per time key using a deterministic minimum-version policy. That is, if multiple supply-demand versions are available for the same time slot, the earliest version is used. This policy is intended to approximate the information that would be available before day-ahead clearing; later revisions are excluded from the main experiment. The selected cache contains 667680 unique supply-demand time keys. Among the selected rows, version 0 accounts for 654377 rows, while higher versions appear only when no version-0 is available for that key.

Each held-out market is evaluated with 83 rolling daily windows. In each LOMO fold, the target market is excluded from adapter training, and the remaining three markets form the source-domain adaptation set. Unless otherwise stated, all feature-interface comparisons share the same regenerated cache and version policy, so observed differences reflect the information interface, not data-preparation inconsistencies.

For the gated LoRA implementation, the source adapter is the same rank-8 LoRA module used in the Source-LoRA baseline, with $\alpha=16$ and insertion into the Chronos-2 q/k/v/o attention projections and output-patch layer. The gate input is the five-dimensional market-state vector $\mathbf{z}_t$ defined in Section~\ref{sec:method}; the gate is linear, $g_\psi(\mathbf{z}_t)=\mathbf{w}^{\top}\mathbf{z}_t+b$, and therefore adds only six trainable parameters. In the gated stage, the pretrained Chronos-2 weights and LoRA matrices are frozen, the reserve-tightness weight is initialized to one while all other gate weights and the bias are initialized to zero, and $\gamma_\psi(\mathbf{z}_t)$ is clipped to $[1.0,3.0]$ with $\beta=2.0$. The gate is trained for 300 source-market steps with batch size 32 and learning rate $10^{-4}$; scale statistics are estimated from source-market training windows only, and no target-market forecast-horizon labels are used.

\subsubsection{Baselines}

Four groups of baselines are considered:
\begin{itemize}
\item Heuristic baselines: previous-day and last-value forecasts;
\item Tree-based methods: XGBoost \cite{chen2016xgboost} and LightGBM \cite{ke2017lightgbm};
\item Deep time-series baselines: PatchTST \cite{nie2023patchtst}, iTransformer \cite{liu2024itransformer}, and the DLinear/NLinear family \cite{zeng2023transformers};
\item Time series foundation models: Chronos-Large \cite{ansari2024chronos}, Moirai-1.1-R-small \cite{woo2024moirai}, and TimesFM-2.5-200M \cite{das2024timesfm}.
\end{itemize}
All reported day-ahead metrics use the same 7-day-to-1-day rolling forecasting protocol. The tree-based and deep-learning models are trained using conventional within-market supervision and serve as reference benchmarks for predictive accuracy, not as target-market-free transfer baselines. The Chronos-2 feature-interface ablations isolate the benefit of exposing richer day-ahead market information under the same regenerated cache.

\begin{table*}[!t]
\centering
\caption{Comparison of the Day-ahead MAE/RMSE of the four-market forecasting protocol}
\label{tab:main_mae}
\begin{threeparttable}
\small
\setlength{\tabcolsep}{2pt}
\begin{tabular}{llrrrrrrrrrr}
\toprule
\multirow{2}{*}{Model} & \multirow{2}{*}{Paradigm} &
\multicolumn{2}{c}{\mGD{}} &
\multicolumn{2}{c}{\mLN{}} &
\multicolumn{2}{c}{\mSD{}} &
\multicolumn{2}{c}{\mSX{}} &
\multicolumn{2}{c}{Average} \\
\cmidrule(lr){3-4}\cmidrule(lr){5-6}\cmidrule(lr){7-8}\cmidrule(lr){9-10}\cmidrule(lr){11-12}
& & MAE & RMSE & MAE & RMSE & MAE & RMSE & MAE & RMSE & MAE & RMSE \\
\midrule
Prev-day naive & baseline & 48.98 & 76.49 & 305.29 & 447.78 & 95.14 & 154.54 & 133.73 & 260.12 & 145.78 & 234.73 \\
Last-value naive & baseline & 68.03 & 104.13 & 257.76 & 424.46 & 113.85 & 185.09 & 133.71 & 228.34 & 143.34 & 235.50 \\
XGBoost & supervised tree & 53.08 & 78.22 & 263.38 & 325.21 & 111.40 & 152.95 & 129.59 & 201.23 & 139.36 & 189.40 \\
LightGBM & supervised tree & 51.82 & 76.76 & 262.95 & 325.70 & 107.96 & 148.45 & 126.69 & 198.99 & 137.35 & 187.47 \\
PatchTST & supervised deep neural network & 49.10 & 71.70 & 267.98 & 345.17 & 82.26 & 121.13 & 118.19 & 193.01 & 129.38 & 182.75 \\
iTransformer & supervised deep neural network & 53.05 & 74.49 & 272.74 & 354.66 & 88.34 & 129.50 & 118.91 & 198.03 & 133.26 & 189.17 \\
DLinear & supervised deep neural network & 42.56 & 63.81 & 243.50 & 318.92 & 87.81 & 123.26 & 123.28 & 197.19 & 124.29 & 175.80 \\
NLinear & supervised deep neural network & 42.66 & 64.55 & 245.98 & 323.98 & 79.47 & 119.36 & 114.96 & 197.08 & 120.77 & 176.24 \\
Chronos-Large & zero-shot foundation model & 42.56 & 66.23 & 257.23 & 419.28 & 102.82 & 168.49 & 115.80 & 210.78 & 129.60 & 216.19 \\
Moirai-1.1-R-small & zero-shot foundation model & 54.22 & 82.37 & 259.84 & 395.32 & 99.62 & 159.89 & 134.86 & 225.14 & 137.13 & 215.68 \\
TimesFM-2.5-200M & zero-shot foundation model & 43.85 & 69.15 & 261.96 & 390.87 & 83.04 & 135.65 & 102.62 & 200.03 & 122.87 & 198.93 \\
Chronos-2 core & zero-shot foundation model & 41.93 & 68.52 & 174.01 & 277.94 & 56.78 & 96.26 & 72.31 & 173.27 & 86.25 & 154.00 \\
Chronos-2 MSMI & zero-shot foundation model & 37.30 & 63.55 & 155.00 & 265.79 & 56.88 & 96.99 & 69.25 & 173.09 & 79.60 & 149.85 \\
Chronos-2 MSMI & Source-LoRA & 36.77 & 62.61 & 147.73 & 243.88 & 55.55 & 94.24 & 67.88 & 170.87 & 76.98 & 142.90 \\
\textbf{Chronos-2 MSMI} & \textbf{Gated LoRA} & \textbf{36.61} & \textbf{62.39} & \textbf{139.24} & \textbf{225.64} & \textbf{54.78} & \textbf{92.61} & \textbf{67.89} & \textbf{170.86} & \textbf{74.63} & \textbf{137.87} \\
\bottomrule
\end{tabular}
\begin{tablenotes}
\footnotesize
\item MAE and RMSE are reported as separate columns on the raw price scale. Average MAE/RMSE are four-market arithmetic means. \mGD{}, \mLN{}, \mSD{}, and \mSX{} denote Guangdong, Liaoning, Shandong, and Shanxi, respectively.
\end{tablenotes}
\end{threeparttable}
\end{table*}

\subsubsection{Information Interfaces}

We define four information interfaces to separate the value of future market information from the value of hand-crafted feature engineering: 
\begin{itemize}
\item The \emph{Core} interface is a compact reference that exposes only the most common day-ahead drivers, including direct load, wind power, and photovoltaic power. 

\item The proposed \emph{MSMI} interface adds raw day-ahead market information, including supply--demand, reserve, maintenance, generator, and intertie fields. 

\item The \emph{Engineered} interface replaces these raw additions with empirical quantities such as net load, renewable share, ramping terms, and reserve margins. 

\item The \emph{Full} interface combines all quantities in the MSMI and engineered interfaces.
\end{itemize}

The latter two interfaces are included solely as diagnostic ablations to assess whether engineered variables can substitute for or complement the original released market fields.

\subsubsection{Metrics}

For point forecasting, we report mean absolute error (MAE) and root mean square error (RMSE) on the raw price scale. For probabilistic forecasting, we report CRPS in \eqref{eq:Appr-CRPS} as the main score, together with PICP and coverage deviation for the central $80\%$ prediction interval. Lower values of MAE, RMSE, CRPS, and coverage deviation indicate better performance, while PICP is evaluated by closeness to the nominal $80\%$ coverage level. Statistical significance is evaluated using one-sided Diebold--Mariano (DM) tests \cite{diebold1995comparing} on daily-window loss differentials from the 83 rolling forecasts per market, with Newey--West Heteroskedasticity and Autocorrelation Consistent (HAC) variance and the Harvey--Leybourne--Newbold small-sample correction \cite{harvey1997testing}. Because the 96 intra-day horizons are aggregated into one daily-window loss, all reported tests use a Newey--West truncation lag of four daily windows. The DM framework is applied to MAE and CRPS score differentials, but not to PICP because coverage is a diagnostic relative to the nominal interval level. Pooled DM tests normalize each market's daily differentials by its mean reference loss or score to avoid dominance by high-price markets.

\subsection{Results}
\label{sec:results}

\subsubsection{Day-ahead forecasting results}

Table~\ref{tab:main_mae} compares MAE and RMSE across the four held-out markets. The MSMI interface reduces the Chronos-2 zero-shot average MAE from 86.25 under the Core interface to 79.60, confirming that known future market information is essential for transferable price forecasting. Source-LoRA further reduces average MAE/RMSE to 76.98/142.90 without target-market labels. The proposed gated LoRA variant gives the lowest average MAE/RMSE of 74.63/137.87, corresponding to a 3.05\% MAE reduction relative to Source-LoRA. The largest improvement appears on \mLN{}, where gated LoRA reduces MAE/RMSE to 139.24/225.64. For other markets, the gated-vs-Source statistical gain is marginal rather than decisive. This result should be interpreted as moderate evidence that reserve-tightness-conditioned calibration can further improve a source-domain adapter in volatile markets.

\begin{figure}[!t]
\centering
\includegraphics[scale=0.6]{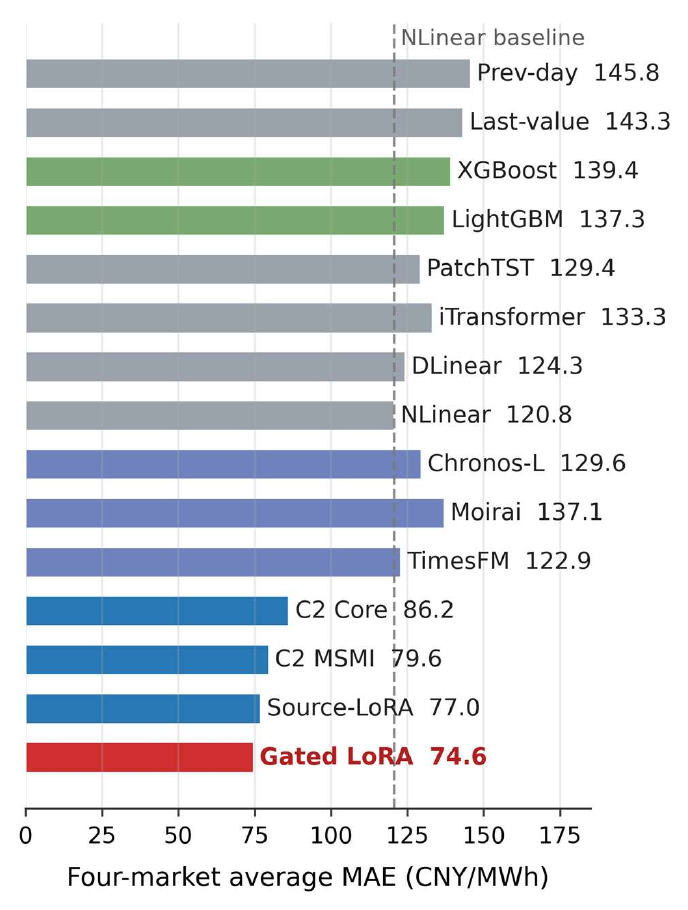}
\caption{Average day-ahead MAE comparison under LOMO evaluation.}
\label{fig:chronos2_main_mae}
\end{figure}

The gate is intentionally small: it freezes the source LoRA matrices and trains only the state-to-scale mapping. This design avoids adding a separate market-specific model while allowing the source adapter to become stronger under tighter operating states. Table~\ref{tab:dynamic_diagnostics} reports the compact control study. A learned global scalar, which removes market-state variation and learns only a task-independent adapter multiplier, remains close to the Source-LoRA reference. Zero and random gate initializations also do not reproduce the main result. The reserve-initialized gate is both more accurate and more stable, although the small matched-vs-shuffled gap means that the causal evidence for exact day-level state pairing should still be interpreted cautiously.

\begin{table}[!t]
\centering
\caption{Gated LoRA robustness and control experiments.}
\label{tab:dynamic_diagnostics}
\begin{threeparttable}
\footnotesize
\setlength{\tabcolsep}{8pt}
\begin{tabular}{lcc}
\toprule
Configuration & Average MAE & CRPS \\
\midrule
Vanilla Source-LoRA & 76.98 & 62.66 \\
Global scalar & $77.001\pm0.000$ & $62.660\pm0.000$ \\
Shuffled state & $74.873\pm0.047$ & $61.402\pm0.049$ \\
Gate, zero init. & $76.962\pm0.032$ & $62.638\pm0.020$ \\
Gate, random init. & $76.601\pm0.531$ & $62.412\pm0.333$ \\
\textbf{Gate, reserve init.} & $\mathbf{74.633\pm0.002}$ & $\mathbf{61.235\pm0.001}$ \\
\bottomrule
\end{tabular}
\begin{tablenotes}
\footnotesize
\item Control rows report mean $\pm$ standard deviation over three gate seeds and four held-out-market averages. The global scalar removes state dependence; the shuffled-state control breaks the pairing between each rolling task and its market-state vector; zero, random, and reserve denote gate initialization. All gated/control rows use source-market training only and no target-market forecast labels.
\end{tablenotes}
\end{threeparttable}
\end{table}

\subsubsection{Statistical significance and robustness}

Table~\ref{tab:dm_tests} reports one-sided Diebold--Mariano tests on daily-window MAE differentials. Each column is written as $\Delta L=L(A)-L(B)$, so a negative statistic favors the first configuration. The MSMI interface significantly improves over Core, Source-LoRA significantly improves over MSMI zero-shot, and gated LoRA improves over Source-LoRA with a marginally significant pooled statistic ($p=0.074$), while \mSX{} remains essentially unchanged. 

\begin{table}[!t]
\centering
\caption{One-sided Diebold--Mariano tests on window-level MAE differentials (83 windows per market).}
\label{tab:dm_tests}
\begin{threeparttable}
\footnotesize
\setlength{\tabcolsep}{8pt}
\begin{tabular}{lccc}
\toprule
Market & MSMI$-$Core & Source$-$MSMI & Gated$-$Source \\
\midrule
\mGD{}  & $-3.78^{***}$ & $-0.87$ & $-1.43^{*}$ \\
\mLN{}  & $-1.83^{**}$  & $-1.32^{*}$ & $-1.19$ \\
\mSD{}  & $0.06$        & $-1.18$ & $-0.91$ \\
\mSX{}  & $-1.21$       & $-1.43^{*}$ & $0.49$ \\
\midrule
Pooled & $-3.15^{***}$ & $-2.24^{**}$ & $-1.45^{*}$ \\
\bottomrule
\end{tabular}
\end{threeparttable}
\par\vspace{1mm}
\begin{minipage}{0.92\columnwidth}
\footnotesize
Each column reports the Harvey–Leybourne–Newbold (HLN)-corrected DM statistic for the stated MAE loss differential $\Delta L=L(A)-L(B)$. Negative statistics favor the first configuration named in the column. The pooled row normalizes each market's daily loss differentials by the reference model's mean loss before concatenation. Significance: $^{*}p<0.10$, $^{**}p<0.05$, $^{***}p<0.01$ (HLN-corrected \cite{harvey1997testing}).
\end{minipage}
\end{table}

As a lightweight robustness check, repeating the source-domain LoRA stage with three random seeds yields four-market average MAEs of 75.32, 77.31, and 76.29 (mean $76.30 \pm 0.99$), and every seed improves upon the zero-shot MSMI level of 79.60. Thus, the source-domain LoRA gain does not rely on a single adapter initialization.

Finally, we verify that the 900-day source-history cap does not limit the reported results. Extending the training pool of the two long-history markets to their full available histories (1419 and 1665 days) changes the four-market average MAE by only $-0.24$, within the seed-level variability. Hence, the Source-LoRA stage saturates before the full multi-year history is consumed, and the reported configuration is not data-starved.

\subsubsection{Probabilistic forecasting results}

Chronos-2 outputs quantile forecasts, allowing probabilistic evaluation without additional generative sampling. Table~\ref{tab:prob_metrics} reports CRPS as the main proper score and PICP@80 as a complementary interval-calibration diagnostic. Source-LoRA improves CRPS from 65.22 to 62.66, and gated LoRA further reduces it to 61.23; the three-seed reserve-gate mean in Table~\ref{tab:dynamic_diagnostics} is nearly identical ($61.235\pm0.001$). The pooled CRPS DM statistic is significant for Source-LoRA over MSMI zero-shot and marginally significant for gated LoRA over Source-LoRA. Pooled PICP@80 decreases from 78.10\% to 73.24\% and then to
70.23\%. Consistently, the macro-average absolute market-wise coverage deviation increases from 5.00 to 6.76 and then to
9.77 percentage points. The market-wise results show that the degradation is not uniform: the gated model exhibits its largest under-coverage on \mLN{}. Thus, gated LoRA improves proper-score performance, but it should not be read as improved interval calibration. A source-validation conformal or interval-scaling layer is still needed for operational use of interval prediction.

% \begin{table}[!t]
% \centering
% \caption{Average probabilistic forecasting metrics.}
% \label{tab:prob_metrics}
% \begin{threeparttable}
% \footnotesize
% \setlength{\tabcolsep}{6pt}
% \begin{tabular}{lccc}
% \toprule
% Metric & MSMI ZS & Source-LoRA & Gated LoRA \\
% \midrule
%    CRPS      &   65.22   &   62.66   &   61.23   \\
% PICP@80 (\%) &   78.10   &   73.24   &   70.23   \\
% $|\mathrm{PICP}-80|$ (p.p.) & 5.00   &  6.76  &  9.77  \\
% \bottomrule
% \end{tabular}
% \begin{tablenotes}
% \footnotesize
% \item p.p.: percentage points. Pooled CRPS DM statistics are $-2.74^{***}$ for Source-LoRA vs. MSMI zero-shot and $-1.59^{*}$ for Gated LoRA vs. Source-LoRA ($^{*}p<0.10$, $^{***}p<0.01$). PICP and coverage deviation are calibration diagnostics and are not tested as losses. 
% \end{tablenotes}
% \end{threeparttable}
% \end{table}

\begin{table*}[!t]
\centering
\caption{Probabilistic forecasting and market-wise central-interval
calibration under LOMO evaluation.}
\label{tab:prob_metrics}
\begin{threeparttable}
\footnotesize
\setlength{\tabcolsep}{15pt}
\begin{tabular}{lccccccc}
\toprule
Method & CRPS & \multicolumn{4}{c}{PICP@80 (\%) by held-out market}
& Pooled PICP@80 & Macro-ACD$_{80}$ \\
\cmidrule(lr){3-6}
& & \mGD{} & \mLN{} & \mSD{} & \mSX{} & (\%) & (p.p.) \\
\midrule
MSMI zero-shot
& 65.22 & 77.87 & 86.21 & 76.02 & 72.31 & 78.10 & 5.00 \\
Source-LoRA
& 62.66 & 74.12 & 75.92 & 73.78 & 69.14 & 73.24 & 6.76 \\
Gated LoRA
& 61.23 & 73.25 & 64.81 & 73.78 & 69.14 & 70.23 & 9.77 \\
\bottomrule
\end{tabular}
\begin{tablenotes}
\footnotesize
\item Pooled PICP@80 is computed after concatenating all forecast
points from the four held-out markets. Macro-ACD$_{80}$ is the
unweighted mean of $|\mathrm{PICP}_{80}^{(m)}-80|$ over the four
markets. Hence, the two summaries use different aggregation orders
and quantify different aspects of interval calibration. p.p.:
percentage points. Pooled CRPS DM statistics are
$-2.74^{***}$ for Source-LoRA versus MSMI zero-shot and
$-1.59^{*}$ for Gated LoRA versus Source-LoRA
($^{*}p<0.10$, $^{***}p<0.01$). PICP and Macro-ACD$_{80}$ are
calibration diagnostics and are not tested as losses.
\end{tablenotes}
\end{threeparttable}
\end{table*}

\subsubsection{Information-interface ablation}

To isolate the value of richer market information, we perform a same-backbone ablation in which Chronos-2 is held fixed and only the input interface changes. Table~\ref{tab:covariate_ablation} reports four zero-shot configurations using the same regenerated data cache, backbone, history length, and rolling evaluation protocol. The MSMI interface obtains the best average MAE among the four zero-shot configurations. It reduces zero-shot MAE from 86.25 under the Core interface to 79.60, a $7.7\%$ relative improvement, and the largest market-level MAE gain appears on \mLN{}. The Engineered interface also improves upon the Core setting, but its smaller gain indicates that hand-crafted market features do not fully substitute for the original supply-demand fields. The Full interface is slightly worse than MSMI on MAE, suggesting that adding every derived variable may introduce redundancy or cross-market inconsistency.

\begin{table}[!t]
\centering
\caption{Market-information interface ablation.}
\label{tab:covariate_ablation}
\begin{threeparttable}
\setlength{\tabcolsep}{8pt}
\begin{tabular}{lccccc}
\toprule
Information & \mGD{} & \mLN{} & \mSD{} & \mSX{} & Average \\
\midrule
Core       & 41.93 & 174.01 & 56.78 & 72.31 & 86.25 \\
MSMI       & \textbf{37.30} & \textbf{155.00} & 56.88 & 69.25 & \textbf{79.60} \\
Engineered & 42.08 & 168.84 & 57.16 & \textbf{67.99} & 84.02 \\
Full       & 38.06 & 157.12 & \textbf{56.23} & 68.13 & 79.88 \\
\bottomrule
\end{tabular}
\begin{tablenotes}
\footnotesize
\item Full denotes the combination of MSMI and engineered variables.
\end{tablenotes}
\end{threeparttable}
\end{table}

\subsubsection{Progressive adaptation spectrum}

\begin{figure}[!t]
\centering
\includegraphics[width=\linewidth]{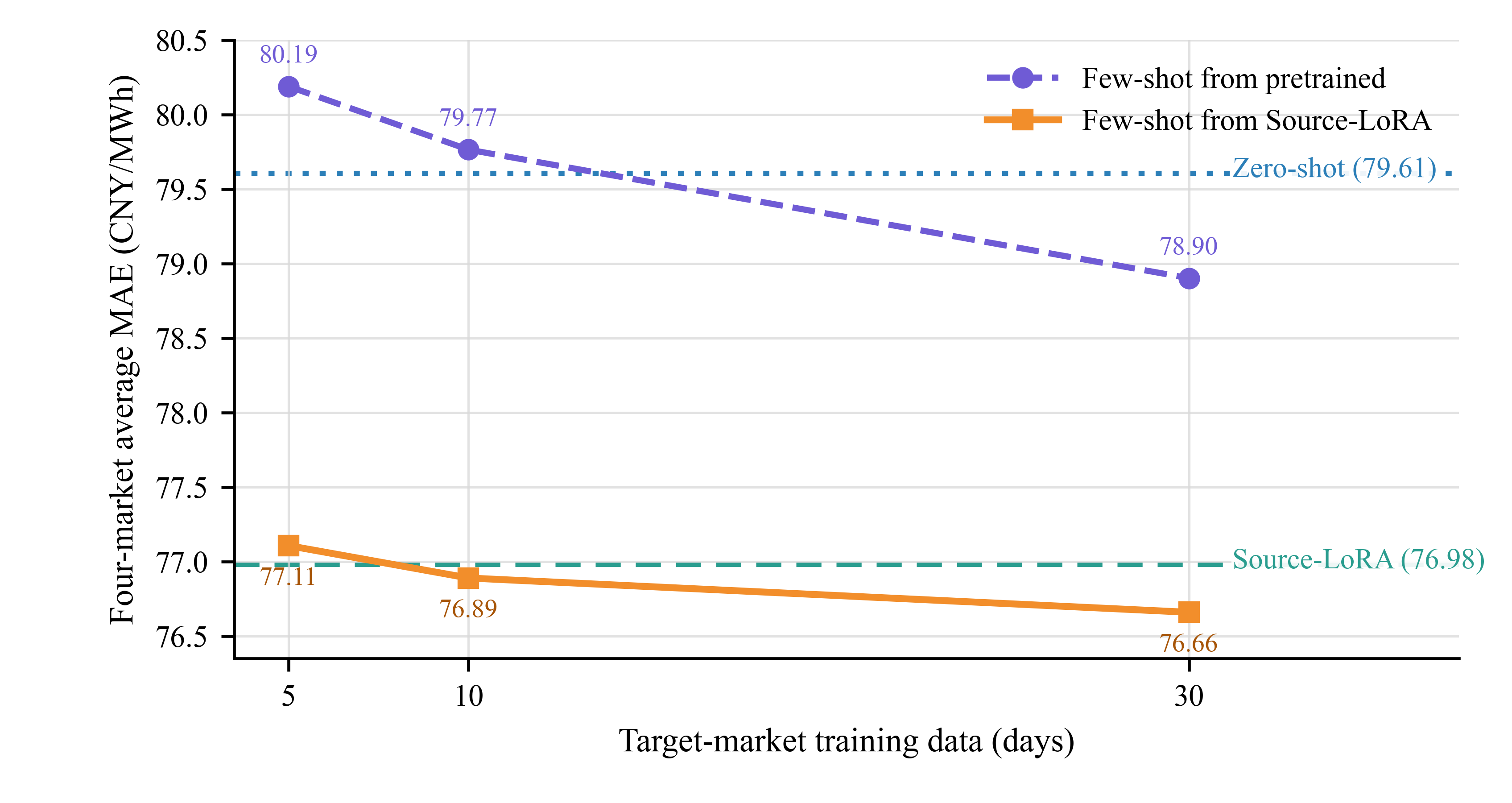}
\caption{Progressive target-market adaptation behavior under the MSMI interface. Points report four-market average MAE computed from the five-seed means in Table~\ref{tab:progressive}. Source-LoRA warm starts provide consistently lower few-shot MAE than direct tuning from pretrained Chronos-2.}
\label{fig:chronos2_progressive}
\end{figure}

Table~\ref{tab:progressive} exhibits the adaptation spectrum from immediate zero-shot deployment to few-shot refinement under the MSMI interface. The few-shot rows report means over five random seeds, and the average day-ahead MAE is illustrated in Fig.~\ref{fig:chronos2_progressive}. Source-domain LoRA reduces the main-run average MAE from 79.60 to 76.98 without using target-market training data, while the three-seed Source-LoRA robustness check gives $76.30 \pm 0.99$. Direct few-shot tuning from pretrained Chronos-2 remains close to the zero-shot level, although its five-seed average improves from 80.19 with 5 target days to 78.90 with 30 target days. Using the source-domain adapter as the warm start gives a consistently stronger few-shot path; the five-seed average decreases from 77.11 with 5 target days to 76.66 with 30 target days. 

\begin{table}[!t]
\centering
\caption{Progressive adaptation spectrum under LOMO evaluation.}
\label{tab:progressive}
\setlength{\tabcolsep}{5pt}
\begin{tabular}{cccccc}
\toprule
Stage & Target data & \mGD{} & \mLN{} & \mSD{} & \mSX{}   \\
\midrule
Zero-shot & None & 37.30 & 155.00 & 56.88 & 69.25  \\
\midrule
Source-LoRA & None & 36.77 & \textbf{147.73} & 55.55 & 67.88  \\
\midrule
Few-shot   &  5 days & 37.39 & 158.64 & 56.01 & 68.72  \\
  from     & 10 days & 37.09 & 157.38 & 55.69 & 68.90  \\
pretrained & 30 days & 37.01 & 154.62 & 55.50 & 68.48  \\
\midrule
Few-shot   &  5 days & 36.85 & 148.93 & 54.98 & \textbf{67.67} \\
 from      & 10 days & 36.60 & 148.67 & 54.83 & 67.46  \\
Source-LoRA& 30 days & \textbf{36.55} & 148.00 & \textbf{54.72} & \textbf{67.37}  \\
\bottomrule
\end{tabular}
\end{table}

\subsubsection{Highly volatile cases}

\begin{figure}[!t]
\centering
\includegraphics[width=\linewidth]{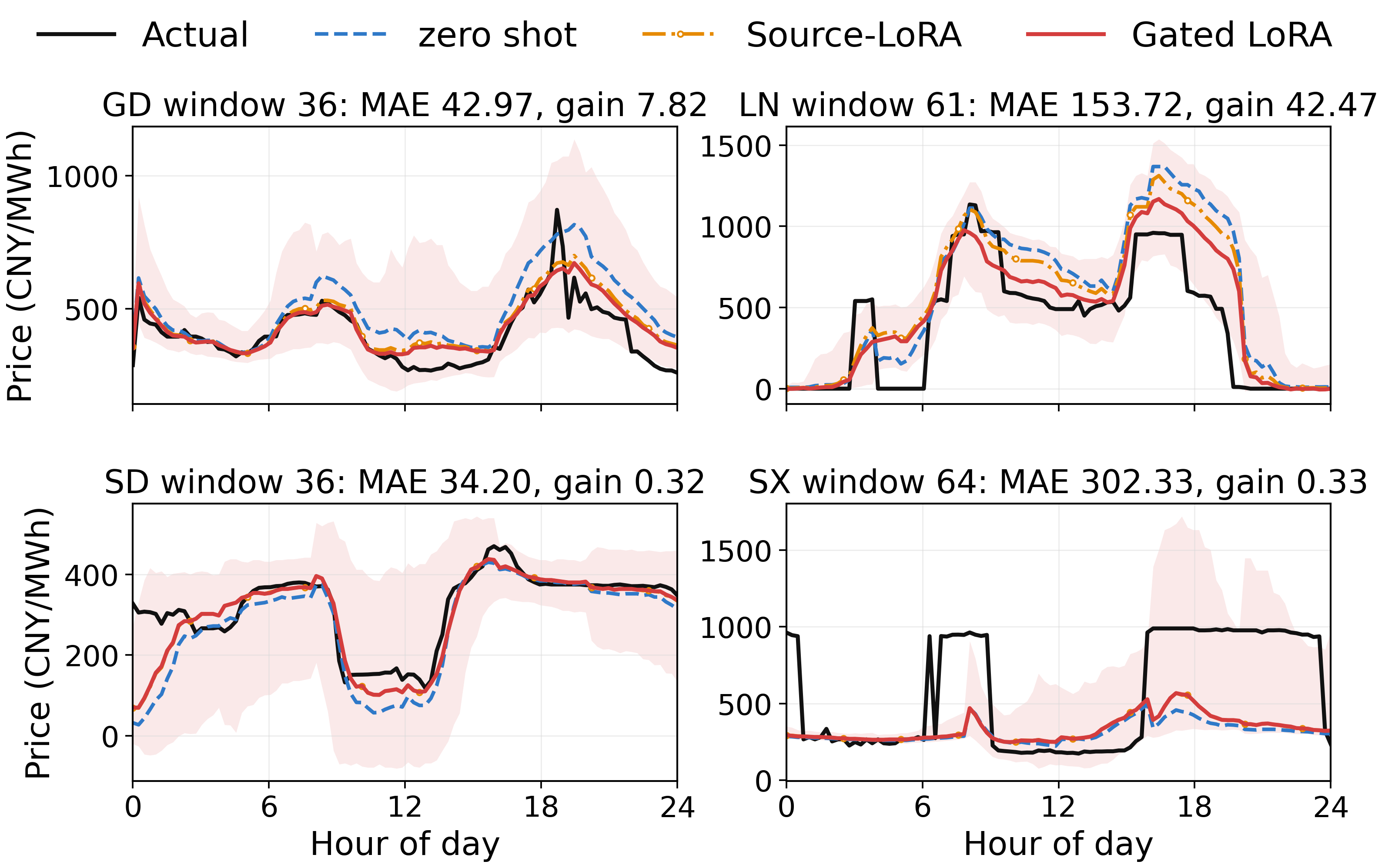}
\caption{Representative day-ahead prediction curves from real Chronos-2 outputs. Each panel compares the actual price, MSMI zero-shot median, Source-LoRA median, gated LoRA median, and gated LoRA 10--90\% quantile band. Source-LoRA is drawn with orange markers to remain visible where it nearly overlaps gated LoRA.}
\label{fig:chronos2_prediction_traces}
\end{figure}

Fig.~\ref{fig:chronos2_prediction_traces} visualizes representative diagnostic windows from the real rolling prediction tasks. The panels compare actual day-ahead prices, MSMI zero-shot medians, vanilla Source-LoRA medians, gated LoRA medians, and the gated LoRA 10--90\% interval. Representative windows are selected by transparent diagnostic rules to cover both high-volatility and lower-error cases across the four markets. This visualization is qualitative; all quantitative conclusions use all 83 rolling windows. The small \mSD{} and \mSX{} gains explain why the Source-LoRA and gated LoRA curves nearly overlap in those panels.

\begin{figure}[!t]
\centering
\includegraphics[width=\linewidth]{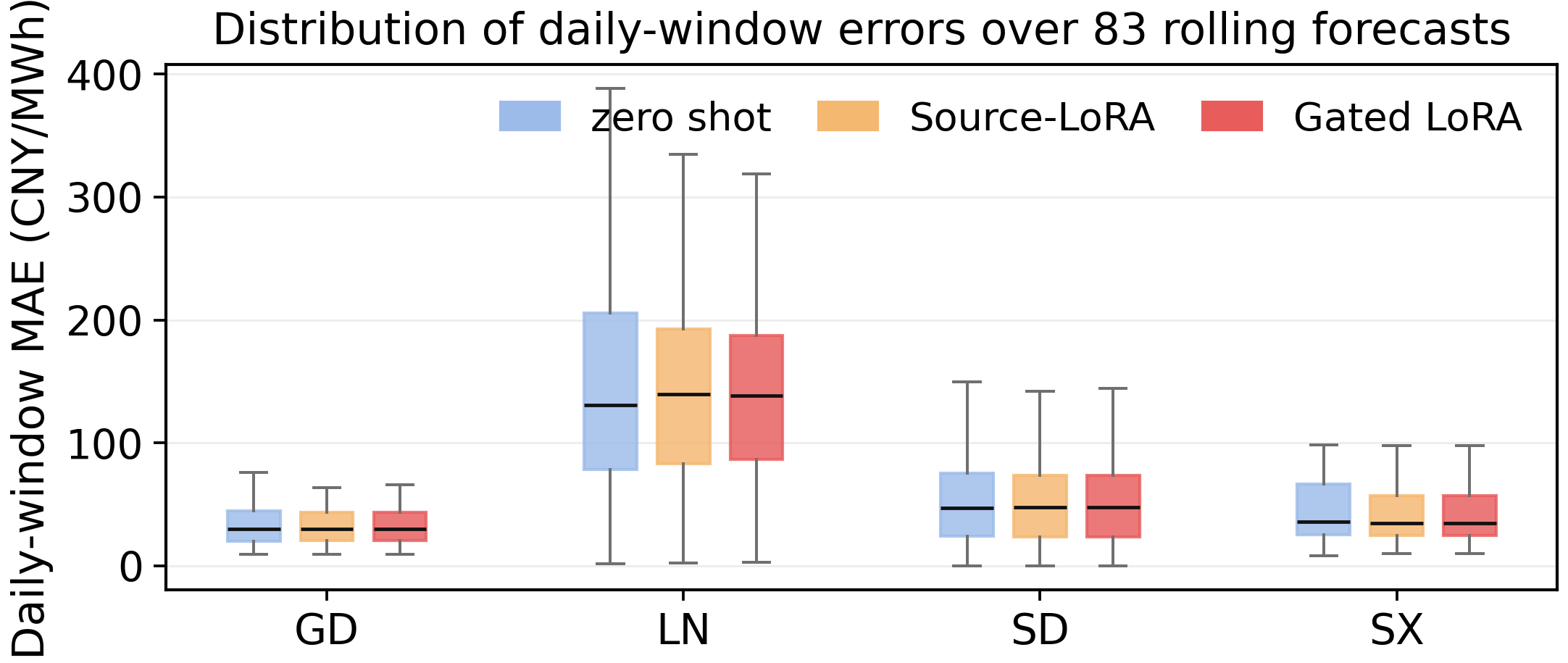}
\caption{Per-window MAE distribution over 83 rolling day-ahead forecasts for each market. Gated LoRA improves the distribution most visibly on \mLN{} and has smaller effects on \mSD{} and \mSX{}, matching the results in Table~\ref{tab:main_mae}.}
\label{fig:chronos2_error_distribution}
\end{figure}

Fig.~\ref{fig:chronos2_error_distribution} further plots the distribution of daily-window MAE over 83 rolling forecasts per market. The boxplots show that gated LoRA most clearly shifts the \mLN{} distribution downward, while \mSD{} and \mSX{} remain close to Source-LoRA. This distributional view is important because average MAE alone can hide whether improvement comes from broad error reduction or from a small number of volatile windows.

\subsection{Discussion}
\label{sec:discussion}

\subsubsection{Why the market-information interface is the dominant source of improvement}

The ablation results in Table~\ref{tab:covariate_ablation} reveal a pattern that is easy to overlook when only aggregate model comparisons are reported: the largest gain comes from giving Chronos-2 access to market information available before day-ahead clearing. This pattern is specific to electricity price forecasting. In wind power forecasting, which originally motivated large time-series model adaptation in energy applications, the target variable is a physical response to meteorological forcing, and historical power alone contains substantial predictive signal \cite{yan2025windltsm}. In day-ahead electricity price forecasting, by contrast, the target is a market-clearing outcome that depends on anticipated supply-demand balance, reserve conditions, generator availability, and intertie schedules. A model that ignores these variables operates with an artificially limited information set, regardless of the size or quality of its pretrained backbone.

The same ablation also shows that market information should not be treated as a generic feature bundle. The Engineered interface improves over Core, but it does not outperform the MSMI fields. This suggests that useful signals are spread across multiple released supply-demand variables. Handcrafted transformations cannot fully replace the original market information. The broader methodological implication is that for structured forecasting tasks driven by known future market conditions, the task interface is a first-class design choice rather than a pre-processing detail.

\subsubsection{Why Gated LoRA helps volatile markets}

The improvement from gated LoRA is most visible on \mLN{}, the most volatile held-out fold. This pattern supports the design motivation: once the MSMI interface supplies the market-clearing information set, the remaining error depends not only on supplied features but also on operating states. A fixed source adapter applies the same correction to ordinary and tight operating windows. The gate instead scales the LoRA update using reserve tightness, net load, renewable share, capacity tightness, and recent price volatility. The control experiments in Table~\ref{tab:dynamic_diagnostics} refine this interpretation: a learned global scalar is not enough, and zero/random gate initializations remain close to the vanilla adapter. The reserve-initialized gate appears to provide a useful economic inductive bias, but the additional gain over Source-LoRA is still moderate. 

\subsubsection{Deployment interpretation}

The results suggest a staged deployment path for markets with limited local history. A new market can start with MSMI zero-shot inference, which already improves substantially over a Core interface. When related-market data are available, Source-LoRA provides a target-label-free alignment step. Gated LoRA can then calibrate this adapter using the current market state and gives the lowest average MAE/RMSE in Table~\ref{tab:main_mae}. Target-market few-shot tuning remains useful as a later-stage local calibration option, although its 30-day five-seed average MAE of 76.66 is only comparable to that of the source-adaptation path and does not demonstrate a clear advantage.

\subsubsection{Limitations and future directions}

Several limitations should be acknowledged. First, future market-information variables are treated as available over the prediction horizon, following the day-ahead information setting used in the protocol. If the market organizer does not reveal all such information, the MSMI interface related benefits may be discounted. Second, the feature-interface ablation identifies useful input groups but does not establish causal effects, since supply, demand, reserve, and generation-availability variables are correlated through system operating conditions. Third, gated LoRA improves MAE and CRPS, but lowers PICP@80 and increases coverage deviation, so interval calibration remains an open issue. The control experiments cover three gate seeds, a learned global scalar, shuffled states, and zero/random/reserve-prior initializations, but they are still limited to four provincial markets. The state gate should be validated on broader markets and longer periods before making a strong general claim. Finally, the present four-market corpus is still modest relative to large-scale foundation-model pretraining datasets, so broader validation across additional comparable day-ahead markets is needed before making claims of universal transferability.

Future work should explore three directions.
\begin{itemize}
\item \textbf{Market-information robustness:} evaluating the same interface under noisy supply--demand forecasts or alternative supply-version policies. 
\item \textbf{Market-adaptive architectures:} extending the proposed market-state gate to dynamic rank allocation or calibrated multi-adapter variants. 
\item \textbf{Spike-aware training objectives:} because average metrics underweight the high-price events that are most consequential for market participants, a spike-weighted loss or quantile-focused calibration stage could improve tail behavior without sacrificing overall accuracy.
\end{itemize}

\section{Conclusion}
\label{sec:conclusion}

This paper investigates market-state-conditioned adaptation of a time-series foundation model for Chinese provincial day-ahead electricity price forecasting. The results lead to three main findings.
\begin{itemize}
\item Task-interface design is central. A same-backbone feature-interface ablation shows that expanding Chronos-2 from the Core load-wind-PV interface to the MSMI interface reduces four-market average zero-shot MAE from 86.25 to 79.60 without changing the pretrained model. This indicates that, for market-clearing targets driven by known future conditions, the foundation-model input interface is not a secondary pre-processing choice but a core modeling decision.

\item Source-domain LoRA provides a lightweight substitute for expensive domain re-pretraining. Under the LOMO protocol, MSMI Chronos-2 reduces average day-ahead MAE to 79.60 in zero-shot mode, and Source-LoRA further reduces it to 76.98 while updating only about $1\%$ of the parameters.

\item Market-state-gated LoRA makes the source adapter responsive to operating conditions and achieves the best average MAE/RMSE of 74.63/137.87. The reserve-initialized gate is stable across three seeds and outperforms global-scalar, shuffled-state, zero-initialized, and randomly initialized controls, while the additional gain over Source-LoRA remains marginally significant in pooled MAE and CRPS DM tests. This supports the view that reserve-tightness-conditioned adapter calibration is a useful and lightweight complement to market-information-aware foundation-model forecasting.
\end{itemize}

Overall, the results support a staged practical workflow: market-information-aware zero-shot inference, source-domain LoRA adaptation, market-state calibration, and optional later target-market refinement. For electricity price forecasting, they also suggest that foundation-model benchmarks should evaluate not only the backbone but also the information interface and adapter-control mechanism.

\bibliographystyle{IEEEtran}
\bibliography{references_chronos2}

@article{weron2014electricity,
  author  = {Weron, Rafa{\l}},
  title   = {Electricity Price Forecasting: A Review of the State-of-the-Art with a Look into the Future},
  journal = {Int. J. Forecasting},
  volume  = {30},
  number  = {4},
  pages   = {1030--1081},
  year    = {2014},
  month   = oct,
  doi     = {10.1016/j.ijforecast.2014.08.008}
}

@article{lago2021forecasting,
  author  = {Lago, Jesus and Marcjasz, Grzegorz and De Schutter, Bart and Weron, Rafa{\l}},
  title   = {Forecasting Day-Ahead Electricity Prices: A Review of State-of-the-Art Algorithms, Best Practices and an Open-Access Benchmark},
  journal = {Appl. Energy},
  volume  = {293},
  pages   = {116983},
  year    = {2021},
  month   = jul,
  doi     = {10.1016/j.apenergy.2021.116983}
}

@article{nowotarski2018recent,
  author  = {Nowotarski, Jakub and Weron, Rafa{\l}},
  title   = {Recent Advances in Electricity Price Forecasting: A Review of Probabilistic Forecasting},
  journal = {Renew. Sustain. Energy Rev.},
  volume  = {81},
  pages   = {1548--1568},
  year    = {2018},
  month   = jan,
  doi     = {10.1016/j.rser.2017.05.234}
}

@article{ziel2018day,
  author  = {Ziel, Florian and Weron, Rafa{\l}},
  title   = {Day-Ahead Electricity Price Forecasting with High-Dimensional Structures: Univariate vs. Multivariate Modeling Frameworks},
  journal = {Energy Econ.},
  volume  = {70},
  pages   = {396--420},
  year    = {2018},
  month   = feb,
  doi     = {10.1016/j.eneco.2017.12.016}
}

@article{guo2020chinamarket,
  author  = {Guo, Hongye and Davidson, Michael R. and Chen, Qixin and Zhang, Da and Jiang, Nan and Xia, Qing and Kang, Chongqing and Zhang, Xiliang},
  title   = {Power Market Reform in {China}: Motivations, Progress, and Recommendations},
  journal = {Energy Policy},
  volume  = {145},
  pages   = {111717},
  year    = {2020},
  month   = oct,
  doi     = {10.1016/j.enpol.2020.111717}
}

@article{gunduz2023transfer,
  author  = {Gunduz, Salih and Ugurlu, Umut and Oksuz, Ilkay},
  title   = {Transfer Learning for Electricity Price Forecasting},
  journal = {Sustain. Energy Grids Netw.},
  volume  = {34},
  pages   = {100996},
  year    = {2023},
  month   = jun,
  doi     = {10.1016/j.segan.2023.100996}
}

@article{dong2023transfer,
  author  = {Dong, Xiaochong and Sun, Yingyun and Dong, Lei and Li, Jian and Li, Yan and Di, Lei},
  title   = {Transferable Wind Power Probabilistic Forecasting Based on Multi-Domain Adversarial Networks},
  journal = {Energy},
  volume  = {285},
  pages   = {129496},
  year    = {2023},
  month   = dec,
  doi     = {10.1016/j.energy.2023.129496}
}

@article{ludwig2015lassoepf,
  author  = {Ludwig, Nicole and Feuerriegel, Stefan and Neumann, Dirk},
  title   = {Putting Big Data Analytics to Work: Feature Selection for Forecasting Electricity Prices Using the {LASSO} and Random Forests},
  journal = {J. Decis. Syst.},
  volume  = {24},
  number  = {1},
  pages   = {19--36},
  year    = {2015},
  month   = jan,
  doi     = {10.1080/12460125.2015.994290}
}

@inproceedings{chen2016xgboost,
  author    = {Chen, Tianqi and Guestrin, Carlos},
  title     = {{XGBoost}: A Scalable Tree Boosting System},
  booktitle = {Proc. 22nd ACM SIGKDD Int. Conf. Knowl. Discovery Data Mining},
  pages     = {785--794},
  year      = {2016},
  month   = aug,
  doi       = {10.1145/2939672.2939785}
}

@inproceedings{ke2017lightgbm,
  author    = {Ke, Guolin and Meng, Qi and Finley, Thomas and Wang, Taifeng and Chen, Wei and Ma, Weidong and Ye, Qiwei and Liu, Tie-Yan},
  title     = {{LightGBM}: A Highly Efficient Gradient Boosting Decision Tree},
  booktitle = {Adv. Neural Inf. Process. Syst.},
  volume    = {30},
  year      = {2017},
  month     = dec
}

@inproceedings{zeng2023transformers,
  author    = {Zeng, Ailing and Chen, Muxi and Zhang, Lei and Xu, Qiang},
  title     = {Are Transformers Effective for Time Series Forecasting?},
  booktitle = {Proc. AAAI Conf. Artif. Intell.},
  volume    = {37},
  number    = {9},
  pages     = {11121--11128},
  year      = {2023},
  month   = jun,
  doi       = {10.1609/aaai.v37i9.26317},
  url       = {https://arxiv.org/abs/2205.13504}
}

@inproceedings{nie2023patchtst,
  author    = {Nie, Yuqi and Nguyen, Nam H. and Sinthong, Phanwadee and Kalagnanam, Jayant},
  title     = {A Time Series Is Worth 64 Words: Long-Term Forecasting with Transformers},
  booktitle = {Proc. Int. Conf. Learn. Represent.},
  year      = {2023},
  url       = {https://arxiv.org/abs/2211.14730}
}

@inproceedings{liu2024itransformer,
  author    = {Liu, Yong and Hu, Tengge and Zhang, Haoran and Wu, Haixu and Wang, Shiyu and Ma, Lintao and Long, Mingsheng},
  title     = {{iTransformer}: Inverted Transformers Are Effective for Time Series Forecasting},
  booktitle = {Proc. Int. Conf. Learn. Represent.},
  year      = {2024},
  url       = {https://arxiv.org/abs/2310.06625}
}

@inproceedings{wu2021autoformer,
  author    = {Wu, Haixu and Xu, Jiehui and Wang, Jianmin and Long, Mingsheng},
  title     = {Autoformer: Decomposition Transformers with Auto-Correlation for Long-Term Series Forecasting},
  booktitle = {Adv. Neural Inf. Process. Syst.},
  volume    = {34},
  pages     = {22419--22430},
  year      = {2021},
  month     = dec
}

@inproceedings{zhou2022fedformer,
  author    = {Zhou, Tian and Ma, Ziqing and Wen, Qingsong and Wang, Xue and Sun, Liang and Jin, Rong},
  title     = {{FEDformer}: Frequency Enhanced Decomposed Transformer for Long-Term Series Forecasting},
  booktitle = {Proc. 39th Int. Conf. Mach. Learn.},
  series    = {Proc. Mach. Learn. Res.},
  volume    = {162},
  pages     = {27268--27286},
  year      = {2022},
  month     = jul,
  publisher = {PMLR},
  url       = {https://arxiv.org/abs/2201.12740}
}

@inproceedings{wang2024timexer,
  author    = {Wang, Yuxuan and Wu, Haixu and Dong, Jiaxiang and Qin, Guo and Zhang, Haoran and Liu, Yong and Qiu, Yunzhong and Wang, Jianmin and Long, Mingsheng},
  title     = {{TimeXer}: Empowering Transformers for Time Series Forecasting with Exogenous Variables},
  booktitle = {Adv. Neural Inf. Process. Syst.},
  volume    = {37},
  pages     = {469--498},
  year      = {2024},
  month   = dec,
  doi       = {10.52202/079017-0015},
  url       = {https://proceedings.neurips.cc/paper_files/paper/2024/hash/0113ef4642264adc2e6924a3cbbdf532-Abstract-Conference.html}
}

@article{lim2021temporal,
  author  = {Lim, Bryan and Arik, Sercan {\"O}. and Loeff, Nicolas and Pfister, Tomas},
  title   = {Temporal Fusion Transformers for Interpretable Multi-Horizon Time Series Forecasting},
  journal = {Int. J. Forecasting},
  volume  = {37},
  number  = {4},
  pages   = {1748--1764},
  year    = {2021},
  month   = oct,
  doi     = {10.1016/j.ijforecast.2021.03.012}
}

@article{rasul2023lagllama,
  author        = {Rasul, Kashif and Ashok, Arjun and Williams, Andrew Robert and Ghonia, Hena and Bhagwatkar, Rishika and Khorasani, Arian and Bayazi, Mohammad Javad Darvishi and Adamopoulos, George and Riachi, Roland and Hassen, Nadhir and Bilo{\v{s}}, Marin and Garg, Sahil and Schneider, Anderson and Chapados, Nicolas and Drouin, Alexandre and Zantedeschi, Valentina and Nevmyvaka, Yuriy and Rish, Irina},
  title         = {Lag-{Llama}: Towards Foundation Models for Probabilistic Time Series Forecasting},
  journal       = {arXiv preprint arXiv:2310.08278},
  year          = {2023},
  eprint        = {2310.08278},
  archivePrefix = {arXiv},
  primaryClass  = {cs.LG},
  url           = {https://arxiv.org/abs/2310.08278}
}

@article{ansari2024chronos,
  author  = {Ansari, Abdul Fatir and Stella, Lorenzo and Turkmen, Ali Caner and Zhang, Xiyuan and Mercado, Pedro and Shen, Huibin and Shchur, Oleksandr and Rangapuram, Syama Sundar and Pineda Arango, Sebastian and Kapoor, Shubham and Zschiegner, Jasper and Maddix, Danielle C. and Wang, Hao and Mahoney, Michael W. and Torkkola, Kari and Wilson, Andrew Gordon and Bohlke-Schneider, Michael and Wang, Bernie},
  title   = {Chronos: Learning the Language of Time Series},
  journal = {Trans. Mach. Learn. Res.},
  year    = {2024},
  month   = nov,
  url     = {https://openreview.net/forum?id=gerNCVqqtR}
}

@inproceedings{woo2024moirai,
  author    = {Woo, Gerald and Liu, Chenghao and Kumar, Akshat and Xiong, Caiming and Savarese, Silvio and Sahoo, Doyen},
  title     = {Unified Training of Universal Time Series Forecasting Transformers},
  booktitle = {Proc. 41st Int. Conf. Mach. Learn.},
  series    = {Proc. Mach. Learn. Res.},
  volume    = {235},
  pages     = {53140--53164},
  year      = {2024},
  month     = jul,
  publisher = {PMLR}
}

@inproceedings{das2024timesfm,
  author    = {Das, Abhimanyu and Kong, Weihao and Sen, Rajat and Zhou, Yichen},
  title     = {A Decoder-Only Foundation Model for Time-Series Forecasting},
  booktitle = {Proc. 41st Int. Conf. Mach. Learn.},
  series    = {Proc. Mach. Learn. Res.},
  volume    = {235},
  pages     = {10148--10167},
  year      = {2024},
  month   = jul,
  publisher = {PMLR},
  url       = {https://proceedings.mlr.press/v235/das24c.html}
}

@article{yu2025pricefm,
  author        = {Yu, Runyao and Gu, Chenhui and Stiasny, Jochen and Wen, Qingsong and Dilov, Wasim Sarwar and Qi, Lianlian and Cremer, Jochen L.},
  title         = {{PriceFM}: Foundation Model for Probabilistic Electricity Price Forecasting},
  journal       = {arXiv preprint arXiv:2508.04875},
  year          = {2025},
  eprint        = {2508.04875},
  archivePrefix = {arXiv},
  primaryClass  = {cs.LG},
  url           = {https://arxiv.org/abs/2508.04875}
}

@inproceedings{hu2022lora,
  author    = {Hu, Edward J. and Shen, Yelong and Wallis, Phillip and Allen-Zhu, Zeyuan and Li, Yuanzhi and Wang, Shean and Wang, Lu and Chen, Weizhu},
  title     = {{LoRA}: Low-Rank Adaptation of Large Language Models},
  booktitle = {Proc. Int. Conf. Learn. Represent.},
  year      = {2022},
  url       = {https://openreview.net/forum?id=nZeVKeeFYf9}
}

@article{gneiting2007proper,
  author  = {Gneiting, Tilmann and Raftery, Adrian E.},
  title   = {Strictly Proper Scoring Rules, Prediction, and Estimation},
  journal = {J. Amer. Statist. Assoc.},
  volume  = {102},
  number  = {477},
  pages   = {359--378},
  year    = {2007},
  month   = mar,
  doi     = {10.1198/016214506000001437}
}

@article{yan2025windltsm,
  author  = {Yan, Jie and Li, Yujia and Wang, Han and Han, Shuang and Shang, Wenlong and Liu, Yongqian},
  title   = {Wind Power Forecasting Based on Large Time Series Model},
  journal = {Engineering},
  year    = {2025},
  month   = nov,
  doi     = {10.1016/j.eng.2025.11.007},
  note    = {Available online 19 Nov. 2025}
}

@article{bottieau2023interpretable,
  author  = {Bottieau, J{\'e}r{\'e}mie and Wang, Yi and De Gr{\`e}ve, Zacharie and Vall{\'e}e, Fran{\c{c}}ois and Toubeau, Jean-Fran{\c{c}}ois},
  title   = {Interpretable Transformer Model for Capturing Regime Switching Effects of Real-Time Electricity Prices},
  journal = {IEEE Trans. Power Syst.},
  volume  = {38},
  number  = {3},
  pages   = {2162--2176},
  year    = {2023},
  month   = may,
  doi     = {10.1109/TPWRS.2022.3195970}
}

@article{zhangfu2024probabilistic,
  author  = {Zhang, Chenxu and Fu, Yong},
  title   = {Probabilistic Electricity Price Forecast With Optimal Prediction Interval},
  journal = {IEEE Trans. Power Syst.},
  volume  = {39},
  number  = {1},
  pages   = {442--452},
  year    = {2024},
  month   = jan,
  doi     = {10.1109/TPWRS.2023.3235193}
}

@article{zhang2022ganlmp,
  author  = {Zhang, Zhongxia and Wu, Meng},
  title   = {Predicting Real-Time Locational Marginal Prices: A {GAN}-Based Approach},
  journal = {IEEE Trans. Power Syst.},
  volume  = {37},
  number  = {2},
  pages   = {1286--1296},
  year    = {2022},
  month   = mar,
  doi     = {10.1109/TPWRS.2021.3106263}
}

@article{uniejewski2018vst,
  author  = {Uniejewski, Bartosz and Weron, Rafa{\l} and Ziel, Florian},
  title   = {Variance Stabilizing Transformations for Electricity Spot Price Forecasting},
  journal = {IEEE Trans. Power Syst.},
  volume  = {33},
  number  = {2},
  pages   = {2219--2229},
  year    = {2018},
  month   = mar,
  doi     = {10.1109/TPWRS.2017.2734563}
}

@article{diebold1995comparing,
  author  = {Diebold, Francis X. and Mariano, Roberto S.},
  title   = {Comparing Predictive Accuracy},
  journal = {J. Bus. Econ. Statist.},
  volume  = {13},
  number  = {3},
  pages   = {253--263},
  year    = {1995},
  month   = jul,
  doi     = {10.1080/07350015.1995.10524599}
}

@article{harvey1997testing,
  author  = {Harvey, David and Leybourne, Stephen and Newbold, Paul},
  title   = {Testing the Equality of Prediction Mean Squared Errors},
  journal = {Int. J. Forecasting},
  volume  = {13},
  number  = {2},
  pages   = {281--291},
  year    = {1997},
  month   = jun,
  doi     = {10.1016/S0169-2070(96)00719-4}
}

@article{chai2019density,
  author  = {Chai, Songjian and Xu, Zhao and Jia, Youwei},
  title   = {Conditional Density Forecast of Electricity Price Based on Ensemble {ELM} and Logistic {EMOS}},
  journal = {IEEE Trans. Smart Grid},
  volume  = {10},
  number  = {3},
  pages   = {3031--3043},
  year    = {2019},
  month   = may,
  doi     = {10.1109/TSG.2018.2817284}
}

@article{jahangir2020rough,
  author  = {Jahangir, Hamidreza and Tayarani, Hanif and Baghali, Sina and Ahmadian, Ali and Elkamel, Ali and Golkar, Masoud Aliakbar and Castilla, Miguel},
  title   = {A Novel Electricity Price Forecasting Approach Based on Dimension Reduction Strategy and Rough Artificial Neural Networks},
  journal = {IEEE Trans. Ind. Informat.},
  volume  = {16},
  number  = {4},
  pages   = {2369--2381},
  year    = {2020},
  month   = apr,
  doi     = {10.1109/TII.2019.2933009}
}

@article{vahedi2026hybrid,
  author  = {Vahedi, Neda and Jolfaei, Alireza and Rehman, Saeed Ur and Ravi, Raja Sekhar},
  title   = {Hybrid Deep Learning Model for Electricity Price Forecasting in Renewable-Dominated Markets: The Case of South Australia},
  journal = {IEEE Trans. Consumer Electron.},
  volume  = {72},
  number  = {1},
  pages   = {1227--1229},
  year    = {2026},
  month   = feb,
  doi     = {10.1109/TCE.2025.3638758}
}

@article{ansari2025chronos2,
  title={Chronos-2: From Univariate to Universal Forecasting},
  author={Abdul Fatir Ansari and Oleksandr Shchur and Jaris Küken and Andreas Auer and Boran Han and Pedro Mercado and Syama Sundar Rangapuram and Huibin Shen and Lorenzo Stella and Xiyuan Zhang and Mononito Goswami and Shubham Kapoor and Danielle C. Maddix and Pablo Guerron and Tony Hu and Junming Yin and Nick Erickson and Prateek Mutalik Desai and Hao Wang and Huzefa Rangwala and George Karypis and Yuyang Wang and Michael Bohlke-Schneider},
  year={2025},
  journal={arXiv preprint arXiv:2510.15821},
  url={https://arxiv.org/abs/2510.15821}
}

\end{document}